\documentclass[letterpaper, 10 pt, conference]{ieeeconf}  

\IEEEoverridecommandlockouts                              

\usepackage{amsmath} 

\usepackage{mathtools}
\usepackage{amssymb}  
\usepackage{bbm}
\usepackage[bb=boondox]{mathalfa}
\usepackage{graphicx} 
\usepackage{hyperref}
\usepackage{booktabs}
\usepackage{threeparttable}
\usepackage{tabularx}
\usepackage{multirow}
\usepackage[caption=false, font=footnotesize]{subfig}
\usepackage{algorithmic}
\usepackage{algorithm}

\usepackage{xcolor}
\usepackage{colortbl}

\usepackage[%
backend=biber,
url=false,
maxbibnames=2,
minbibnames=1,
bibstyle=ieee,
]{biblatex}
\AtBeginBibliography{\footnotesize}
\AtEveryBibitem{
  \clearfield{doi}
  \clearfield{isbn}
  \clearfield{issn}
  \clearfield{month}
  \clearlist{publisher}
  \ifentrytype{inproceedings}{
    \clearfield{arxivId}
    \clearfield{archivePrefix}
    \clearfield{eprint}
  }{}
  \ifentrytype{article}{
    \clearfield{arxivId}
    \clearfield{archivePrefix}
    \clearfield{eprint}
  }{}
  \ifentrytype{report}{
  }{}
}
\renewbibmacro{in:}{%
  \ifentrytype{inproceedings}{%
    \setunit{}
    \addperiod\addspace In \textit{Proc.\ of the}}%
  {\printtext{\bibstring{in}\intitlepunct}}
}
\DeclareSourcemap{
  \maps{
    \map{ 
      \step[fieldsource=url,
      match=\regexp{\{\\\_\}|\{\_\}|\\\_},
      replace=\regexp{\_}]
    }
    \map{ 
      \step[fieldsource=url,
      match=\regexp{\{\$\\sim\$\}|\{\~\}|\$\\sim\$},
      replace=\regexp{\~}]
    }
    \map{ 
      \step[fieldsource=url,
      match=\regexp{\{\\\x{26}\}},
      replace=\regexp{\x{26}}]
    }
  }
}
\title{\LARGE \bf
Multi-Source Soft Pseudo-Label Learning with 
\\Domain Similarity-based Weighting for Semantic Segmentation
}

\author{Shigemichi Matsuzaki$^{1}$, Hiroaki Masuzawa$^{1}$, and Jun Miura$^{1}$
\thanks{$^{1}$S. Matsuzaki, H. Masuzawa, and J. Miura are 
with Department of Computer Science and Engineering, Toyohashi University of Technology,
Hibarigaoka 1-1, Tenpaku-cho, Toyohashi, Aichi, Japan {\tt\small matsuzaki@aisl.cs.tut.ac.jp}}%
\thanks{Source code: \url{https://github.com/ShigemichiMatsuzaki/MS2PL}}
}

\begin{document}

\maketitle
\thispagestyle{empty}
\pagestyle{empty}


\begin{abstract}
	This paper describes a method of domain adaptive training for semantic segmentation
	using multiple source datasets that are not necessarily relevant to the target dataset.
	We propose a \textit{soft} pseudo-label generation method by integrating
	predicted object probabilities from multiple source models.
	The prediction of each source model is weighted
	based on the estimated domain similarity between the source and the target datasets
	to emphasize contribution of a model trained on
	a source that is more similar to the target and
	generate reasonable pseudo-labels.
	We also propose a training method
	using the soft pseudo-labels considering
	their entropy to
	fully exploit information from the source datasets
	while suppressing the influence of
	possibly misclassified pixels.
	The experiments show comparative or better performance
	than our previous work and
	another existing multi-source domain adaptation method,
	and applicability to a variety of target environments.
\end{abstract}


\section{Introduction}

Semantic segmentation based on deep neural networks (DNNs)
has been used as a common and strong tool for
scene recognition of autonomous mobile agents.
However, efficiency of training remains a critical problem.
Usually, networks are trained on
a large amount of manually labeled data collected via
a laborious annotation process.
Although rich datasets are publicly available
for some actively studied scenes, e.g., urban scenes
where autonomous driving is a hot topic \cite{Cordts2016,Brostow2009},
there are way fewer datasets for specific scenes,
such as greenhouses and unstructured scenes.

Domain adaptation (DA) is a task to adapt
a model pre-trained on a source dataset
to a target dataset,
and unsupervised domain adaptation (UDA) is a problem setting
where the target dataset has no ground truth labels.
Especially in autonomous driving community, UDA for semantic segmentation
has been actively studied as a promising approach
to efficient training by exploiting simulated photorealistic images
generated by video games \cite{Richter2016} 
or dedicated simulators \cite{Ros2016,Dosovitskiy2017}. 
However, again, it is not easy to collect suitable
source datasets for many other environments.

\begin{figure}[tb]
	\centering
	\includegraphics[width=0.85\hsize]{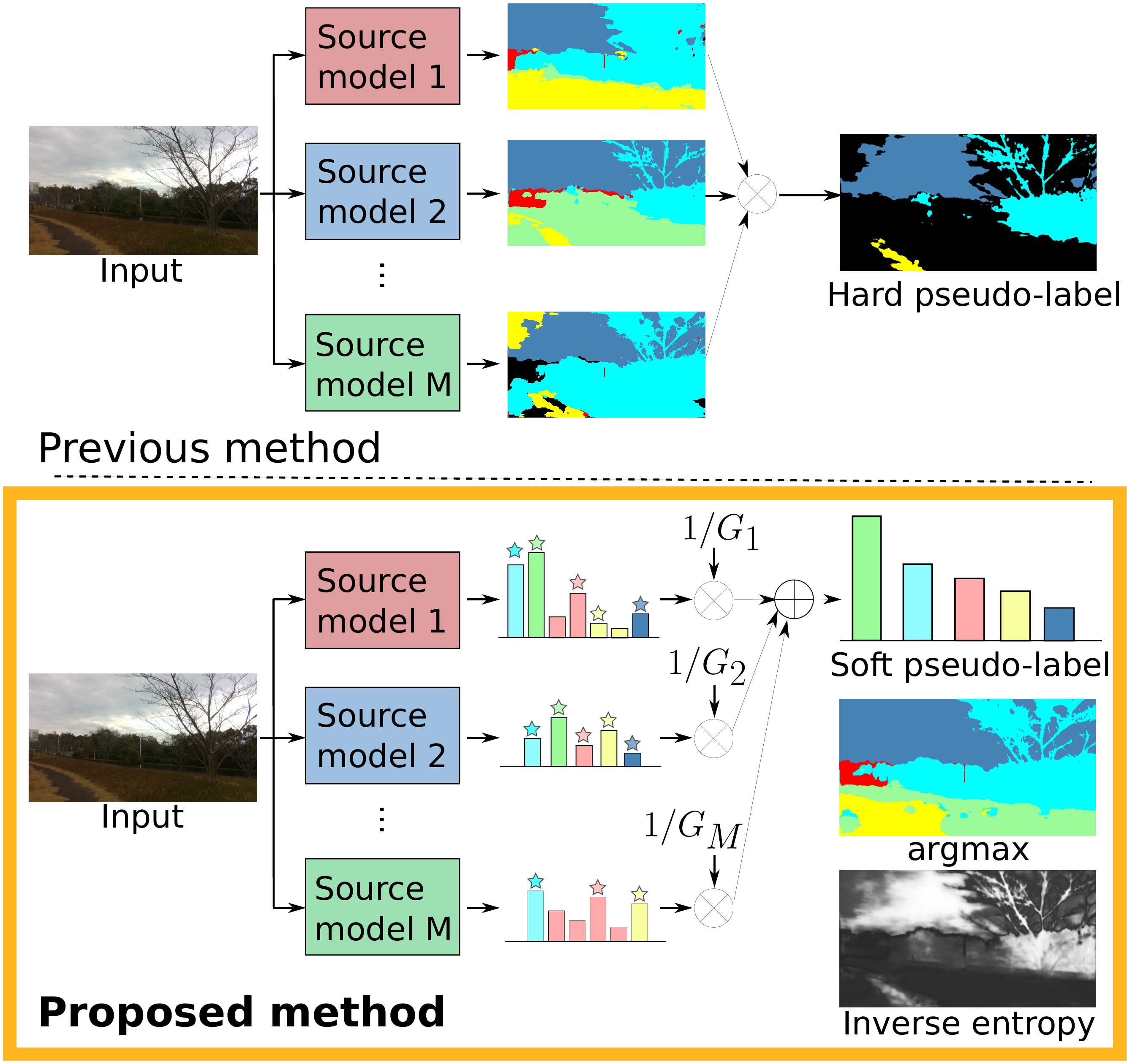}
	\vspace{-8pt}
	\caption{\textbf{Top}: Previous method \cite{Matsuzaki2022a}
		generates pseudo-labels using
		multiple source models. A label is assigned only
		if all models agree with each other to remove wrong labels.
		If a class is not included in a source, the class never appears
		in the resulting pseudo-labels.
		\textbf{Bottom}: Proposed method. It generates \textit{soft} pseudo-labels
		by summing predicted scores weighted
		by inverse domain gap, i.e., domain similarity.
		The degree of agreement of the source models are represented by
		the inverse entropy of the soft pseudo-labels.
		It can also involve labels not present in some source datasets.}
	\label{fig:introduction}
\end{figure}

In our previous work \cite{Matsuzaki2022a},
we proposed a method to train a semantic segmentation model
for greenhouse images
using multiple publicly available datasets
of scenes not relevant but related to greenhouses,
e.g., urban scenes and unstructured outdoor scenes,
as source datasets to overcome the aforementioned problem.
The method utilizes source datasets
with different scenes and label sets to train a model on
the target dataset.
The outputs from each source model are
converted to pixel-wise one-hot labels in the common target label sets
and merged based on unanimity of all the models
(see Fig. \ref{fig:introduction}).
This allows reliable labels to be effectively extracted.

While such a strict criterion 
contributes to excluding possibly wrong labels
and results in good performance, 
it completely ignores information
of many pixels excluded from the training, leading to sparsity of valid labels.
A limitation of the method stemming from this problem is
that the source datasets must have at least one corresponding object class
for all target classes
to enable unanimity-based label selection.
This restricts the choice of source datasets.

In this paper, we extend our previous work \cite{Matsuzaki2022a} and
propose a novel multi-source pseudo-label generation method.
Instead of selecting one-hot pseudo-labels
based on unanimity,
we generate \textit{soft} pseudo-labels which take a form of
class probability distribution on each pixel by integrating
the outputs from the source models.
To integrate the outputs from the source models,
we take into account
quantitatively evaluated domain similarities
between the target data and each source dataset 
so that the predictions of a model trained on
a dataset closer to the target is emphasized more.
In training, we weight the loss values of each pixel
with a value inversely proportional to the entropy
of the soft pseudo-label.

The contributions of the paper are as follows:
\begin{enumerate}
	\item A soft pseudo-label generation method using multiple source datasets considering the domain gap between the source datasets and the target dataset
	\item A training method using the soft pseudo-labels considering the entropy of the labels
\end{enumerate}

\section{Related Work}

\subsection{Domain adaptation for semantic segmentation}

DA is attracting attention as a method to
workaround the necessity of manually annotating
the dataset of the target task.
Specifically in training of DNNs for semantic segmentation,
there are two major approaches:
\textit{domain alignment} and
\textit{pseudo-label learning} \cite{Zou2018,Zou2019}.
The former is realized via minimizing divergence metrics
\cite{Saito2018a},
adversarial learning \cite{Hoffman2018,Vu2019}, etc.
These two approaches are not mutually exclusive and
jointly used in many methods \cite{Zheng2021,Zhang2021}.

Multi-source Domain Adaptation (MDA) for semantic segmentation
has been actively studied in the last few years.
Zhao et al. \cite{Zhao2019a} pioneered
an MDA method for semantic segmentation
by extending the work by \cite{Hoffman2018}
to the multi-source setting.
He et al. \cite{He2021} proposed 
to use multiple collaboratively trained source models
to generate pseudo-labels by
simply summing their predicted probabilities.
This approach assumes that
the source datasets are equally similar
to the target dataset, which is not the case
in our problem setting.
In our previous work \cite{Matsuzaki2022a},
we proposed a multi-source pseudo label learning method
specifically for training a model on
greenhouse images leveraging multiple publicly available datasets.
Unlike \cite{He2021}, our previous method
showed effectiveness on transferring knowledge
from source datasets structurally dissimilar to
the target dataset.
However, the pseudo-labels generated in \cite{Matsuzaki2022a}
are based on unanimity of the source models,
which inherently discards a lot of pixels
and is incapable of involving labels only
in some of the source datasets.
Moreover, it does not consider similarities of the source datasets
with the target.
The present work is an attempt to resolve the limitation
of the previous work and to gain better applicability
of the method.

\subsection{Domain gap evaluation}

The domain gap, or domain shift,
generally stems from the discrepancy between
data distributions of the two data domains.
There have been several metrics to
measure the shift between two domains,
such as
Kulback-Leibler divergence (KLD),
Maximum Mean Discrepancy (MMD) \cite{Gretton2012},
and $\mathcal{H}\Delta\mathcal{H}$-divergence \cite{Ben-David2010c}.
These metrics are often used as an objective to
minimize in domain adaptive tasks
to learn domain-invariant knowledge
\cite{Long2015a,Long2016}.

Liu et al. \cite{Liu2021} proposed a data-driven method of
domain gap evaluation.
While the discrepancy metrics such as MMD
are used as an objective of minimization during adaptation,
the method focuses on
evaluating the discrepancy between the source
and the target dataset themselves.
We employ this method to evaluate
relative domain gaps between different source datasets
against the target dataset.

\section{Preliminaries}

\subsection{Notations}

Formally, we assume $M$ labeled source datasets $S_1,\cdots,S_M$ and
an unlabeled target dataset $S_T$.
A source dataset $S_i$ is  a set of $N_i$ input images $X_i=\{x_{i,j}\}^{N_i}_{j=1}$
and corresponding pixel-wise semantic label maps $Y_i=\{y_{i,j}\}^{N_i}_{j=1}$
with $C_i$ classes.
The target dataset $S_T$ consists of a set of $N_T$ unlabeled images $X_T=\{x_{T,j}\}^{N_T}_{j=1}$.
Let $F\left(\cdot; \theta_k\right)$ denote a segmentation model with learnable weights $\theta_k$
trained on a source dataset $S_k$.
In addition, let $\prescript{k}{}{p}_{i,j}\in \mathbb{R}^{ H\times W\times C_k}$ and
${}^kf_{i,j}\in\mathbb{R}^{H\times W \times D}$
denote a pixel-wise object probability and $D$-dimensional intermediate features
produced by $F\left(x_{i,j}; \theta_k\right)$, respectively.
For a tensor $z\in \mathbb{R}^{H\times W\times C}$,
$z^{\left(h,w,c\right)}$ denotes an element at
index $\left(h, w, c\right)$.

\subsection{Network architecture}
The proposed method does not rely on a specific network architecture.
It, however, assumes two parallel segmentation decoders,
namely \textit{main} and \textit{auxiliary} branches,
to enable uncertainty-based loss rectification \cite{Zheng2021,Matsuzaki2022a}.
Let $\prescript{k}{m}p_{i, j}$ and
$\prescript{k}{a}p_{i, j}$ denote
object scores predicted by the main branch
and the auxiliary branch, respectively.
The object probability $\prescript{k}{}{p}_{i,j}$
is specifically given as follows \cite{Zheng2021}:
\begin{equation}
	\prescript{k}{}{p}_{i,j} = \text{Softmax}\left(\prescript{k}{m}p_{i, j} + w\prescript{k}{a}p_{i, j}\right)%
\end{equation}
where $w$ is a weight parameter and is set to $0.5$ as in \cite{Zheng2021}.

%
%
%


\section{Multi-source soft pseudo-label generation \\ considering domain similarity}
\label{sec:pseudo_label_generation}


In this section, we describe the method of
generating soft pseudo-labels
utilizing segmentation models pre-trained on
the source datasets
as the first step of the proposed method.

\subsection{Pseudo-label generation}

First, we train semantic segmentation models
using the source datasets.
Following \cite{Matsuzaki2022a},
the models are trained using the ordinary cross entropy loss:
\begin{equation}
	L_{ce}\left(p, y\right)^{(h,w)}=-\sum_{c\in C}
	y^{\left(h,w,c\right)}\log p^{\left(h,w,c\right)},
\end{equation}
where $p$ denotes the predicted pixel-wise probability distributions,
and $y$ denotes a one-hot label map.



After pre-training,
we generate pseudo-labels by integrating outputs on the target images
from the source models.
The outputs are weighted with domain similarity,
i.e., inverse of domain gap
between the source and the target datasets.
To evaluate relative domain gap,
we employ a method by Liu et al. \cite{Liu2021}
that uses the entropy of predictions
as a measure of the domain gap.
The domain gap $\prescript{i}{}{G}_j$ between source dataset $S_i$ and
a target image $x_{T,j}$ is calculated as follows:
\begin{equation}
	\prescript{i}{}{G}_j = \frac{1}{\log{C_i}}E\left({}^ip_{T,j}\right),
\end{equation}
where $E\left(\cdot\right)$ is the entropy of the prediction defined as follows:
\begin{equation}
	E\left({}^ip_{T,j}\right) =-\sum_{h,w}\sum^{C_{i}}_{c=1}{}^ip_{T,j}^{(h,w,c)}\log\left({}^ip_{T,j}^{(h,w,c)}\right).
\end{equation}
The larger $\prescript{i}{}{G}_j$ is,
the farther the source dataset $S_i$ and the target image $x_{T,j}$ are.
This method is based on an observation that
the large domain gap makes fuzzy classification scores \cite{Liu2021}
and leads to a high entropy of the distribution.
Since the value range of entropy depends on the number of classes,
the value is normalized by the maximum value, i.e., $\log{C_i}$.

\begin{figure}[tb]
	\centering
	\includegraphics[width=0.65\hsize]{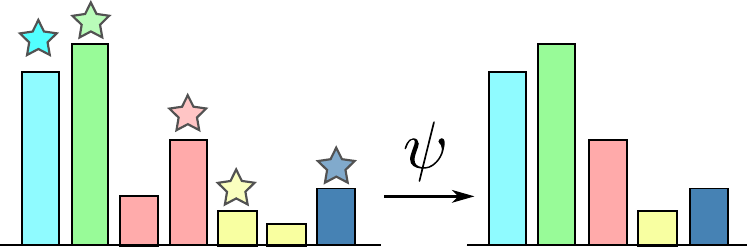}
	\caption{Label conversion function $\psi$. The colors of the bars
		represent the corresponding target labels.
		For each target class,
		the highest score among a group of corresponding source classes
		is selected.  The scores are then normalized to form
		a probability distribution.}
	\label{fig:label_conversion}
\end{figure}

Using the estimated domain gaps, soft pseudo-labels are generated as follows:
\begin{equation}
	\hat{y}_{j}^{(h,w)} =\text{Softmax}\left(\sum^{M}_{i=1} \frac{1}{\prescript{i}{}{G}_j}\psi_i\left({}^{i}p_{T,j}^{(h,w)}\right)\right),
\end{equation}
where $\psi_i: \mathbb{R}^{C_i}\rightarrow \mathbb{R}^{C_T}$ denotes
a function to convert a probability distribution in the source label space
to the target label space (see Fig. \ref{fig:label_conversion}).
The label mapping is heuristically defined for each source
as in \cite{Matsuzaki2022a}.

\subsection{Analysis of the soft pseudo-labels}
\label{seq:pseudo-label_analysis}

\begin{figure*}[tb]
	\raggedleft
	\begin{minipage}{0.4\hsize}
		\centering
		\includegraphics[width=\hsize]{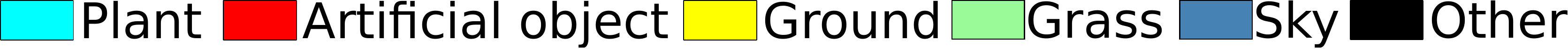}
	\end{minipage} \\
	\centering
	\begin{minipage}{0.90\hsize}
		\includegraphics[width=\hsize]{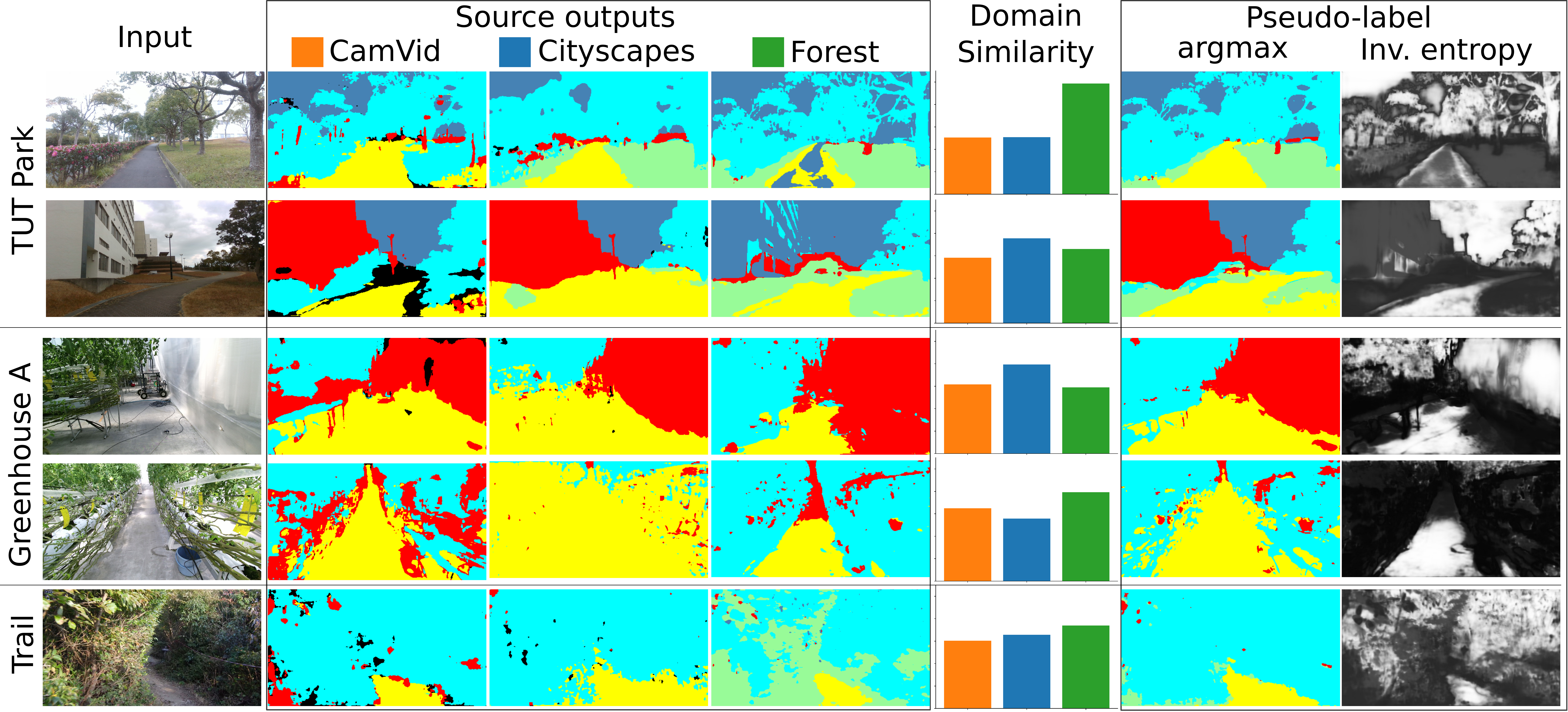}
	\end{minipage}
	\vspace{-3pt}
	\caption{Predictions of the source models, domain similarity (inverse domain gap),
		and pseudo-labels (labels with the maximum probability, and inverse of label entropy).
		We used CamVid \cite{Brostow2009}, Cityscapes \cite{Cordts2016},
		and Freiburg Forest \cite{Valada2017} as source datasets
		(descriptions about the datasets are in Table \ref{table:datasets}).
		In inverse entropy images, a darker pixel indicates a lower weight, i.e., higher entropy.
	}
	\label{fig:pseudo_label_examples}
\end{figure*}

Fig. \ref{fig:pseudo_label_examples} shows examples of
predictions from each source model,
and the predicted relative domain similarity
(the inverse of domain gap) of the inputs,
as well as argmax and inverse label entropy of the generated soft pseudo-labels.
The first image of TUT Park dataset
having a large area of \textit{plant} and \textit{grass} on both sides
was evaluated similar to the Forest dataset,
which shares a similar structure.
In the second image, on the other hand,
there is a large building,
which is seen more often in urban scenes,
and it resulted in higher similarity with Cityscapes.
In a case where the similarity between
the sources and the target is difficult to evaluate
due to large difference
(e.g., greenhouse vs. urban / outdoor scenes),
the domain similarity serves as relative confidence measures.
For example, in the second example of Greenhouse A,
the predictions by CamVid and Forest models are fairly reasonable
and the estimated similarities are relatively high.
Conversely, the Cityscapes model
performs poorly and exhibits lower similarity.
From the observation above, it seems reasonable
to use the domain similarity scores
as importance weights in pseudo-label generation.

By merging the source predictions using the proposed method described above,
fairly accurate pseudo-labels are generated
(as can be seen in the column of ``argmax'').
In addition, entropy tends to be higher (a weight is lower) on pixels
on which all source models agree with each other.
This observation leads to our training method
utilizing the inverse entropy of the soft pseudo-labels,
which is a soft alternative of
the unanimity-based label selection \cite{Matsuzaki2022a}.

Notably in the examples of \textit{TUT Park} dataset,
the \textit{grass} class is assigned in the pseudo-labels
though it is not present in CamVid dataset.
In the previous method \cite{Matsuzaki2022a},
such flexible label assignment is not possible
because of the strict unanimity-based label selection.
This characteristic allows for more flexible choice of source datasets.

\section{Network training}

We train a target model
using the soft pseudo-labels with
the loss function specifically tailored for
the soft pseudo-labels.
We also employ an existing method \cite{Zhang2021}
for training robust to misclassified pseudo-labels.

\subsection{Loss function}

As a base classification loss,
we employ symmetric cross-entropy (SCE) loss \cite{Wang2019e},
a variant of cross-entropy loss robust to noisy labels, following \cite{Zheng2021}:
\begin{eqnarray}
	L_{sce}^{\left(h, w, c\right)} = \alpha L_{ce}\left(\prescript{T}{}p^{\left(h, w, c\right)}_{T, j}, \delta\left(\hat{y}_{j}^{(h,w)}\right)\right) \nonumber \\
	+  \beta L_{ce}\left(\delta\left(\hat{y}_{j}^{(h,w)}\right), \prescript{T}{}p^{\left(h, w, c\right)}_{T, j}\right),
\end{eqnarray}
where $\alpha$ and $\beta$ are balancing parameters
and set to $0.1$ and $1.0$, respectively.
$\delta\left(\cdot\right)$ denotes a function to convert
a soft label to one-hot label.
In implementation,
the one-hot label is clamped to $[1e-4, 1.0]$
to avoid numerical error \cite{Zhang2021}.

Based on the observation in \ref{seq:pseudo-label_analysis},
we consider weighting loss values on each pixel
with the inverse value of label entropy.
The pixel-wise weight is calculated as follows:
\begin{equation}
	W^{(h,w)}_{j}=\exp{\left(-\lambda_{scale} \cdot E\left(\hat{y}^{(h,w)}_{j}\right)\right)},
	\label{eq:label_entropy_weight}
\end{equation}
where $\lambda_{scale}$ is a scaling parameter.
Using eq. \eqref{eq:label_entropy_weight},
we calculate a weighted SCE loss:
\begin{equation}
	L_{w\_sce}^{(h,w)} = W^{(h,w)}_j\cdot L_{sce}^{(h,w)}.
	\label{eq:weighted_sce}
\end{equation}
This way, loss values on the pixels with low label entropy
are weighted more, and it results in a similar effect to the previous
unanimity-based \textit{hard} pseudo-labels.

Following \cite{Matsuzaki2022a},
we also introduce a loss rectification method based on
the pixel-wise uncertainty of the prediction proposed by \cite{Zheng2021}
to suppress the effect of pixels with high uncertainty,
whose pseudo-labels are more likely to be wrong.
The uncertainty is estimated as
Kulback-Leibler divergence
between the branches on each pixel as follows:
\begin{equation}
	\label{eq:kld}
	L_{kld}^{\left(h, w\right)} = \sum_{c\in C} \prescript{T}{m}p^{\left(h, w, c\right)}_{T, j}
	\log{\frac{\prescript{T}{m}p^{\left(h, w, c\right)}_{T, j}}{\prescript{T}{a}p^{\left(h, w, c\right)}_{T, j}}}.
\end{equation}
The rectified classification loss is defined as follows:
\begin{equation}
	L_{rect}^{(h,w)}=\exp{\left(-L_{kld}^{\left(h, w\right)}\right)}L_{w\_sce}^{\left(h, w\right)}.
\end{equation}

Along with the classification loss,
we add the entropy loss of the predictions to
force the network to clearly distinguish the classes,
as used e.g., in \cite{Vu2019}:
\begin{equation}
	L_{ent}^{(h,w)}=E\left(\prescript{T}{}p^{(h,w)}_{T, j}\right),
\end{equation}
The overall loss is as follows:
\begin{equation}
	L_{all}=\sum_{h,w}\left(L_{rect}^{(h,w)} + \lambda_{ent} L_{ent}^{(h,w)} + \lambda_{kld}L_{kld}^{(h,w)}\right),
\end{equation}
where $\lambda_{ent}$ and $\lambda_{kld}$ denote balancing parameters.

\subsection{Prototype-based pseudo-label rectification}

Along with the soft pseudo-labels,
we employ ProDA \cite{Zhang2021}
to make the training process robust to noisy pseudo-labels.
The core of the method is to
save the source model's prediction as
soft pseudo-labels $\{\hat{y}_{j}\}_{j=1}^{N_T}$
and rectify them during training.
The labels are rectified based on
the distance between the pixel feature and
the \textit{prototype},
which is the representative feature for each class.
See \cite{Zhang2021} for details.


The feature-wise weights are calculated as follows:
\begin{equation}
	\omega^{(h,w,c)}_{j}=\frac{\exp \left(-\left\|\prescript{T}{}{\tilde{f}}^{(h,w)}_{j}-\eta^{(c)}\right\| / \tau\right)}{\sum_{c^{\prime}} \exp \left(-\left\|\prescript{T}{}{\tilde{f}}^{(h,w)}_{j}-\eta^{\left(c^{\prime}\right)}\right\| / \tau\right)},
\end{equation}
where $\eta^{(c)}$ denotes the prototype for class $c$,
initialized as a mean of the features predicted as $c$
and updated during training.
$\left\|\cdot\right\|$ denotes the Euclidean norm,
and $\tau$ denotes a temperature parameter
that controls the degree of bias of the distribution.
$\prescript{T}{}{\tilde{f}}_{j}$ is a feature vector
from a momentum encoder \cite{He2020},
i.e., a model identical to $F\left(\cdot;\theta_T\right)$
whose parameters are updated via exponential moving average (EMA).

Using $\omega_j$, the soft pseudo-labels are rectified as follows:
\begin{equation}
	\hat{y}'^{(h,w,c)}_{j} = \frac{\omega^{(h,w,c)}_{j} \hat{y}^{(h,w,c)}_{j}}{\sum_{c^{\prime}}\omega^{(h,w,c^{\prime})}_{j} \hat{y}^{(h,w,c^{\prime})}_{j}}\
	\label{eq:label_rectification}
\end{equation}
In our method, the soft pseudo-labels are generated
by the method described in \ref{sec:pseudo_label_generation}
and rectified by eq. \eqref{eq:label_rectification}
in training.

\section{Experiments}

\subsection{Experimental setting}

\subsubsection{Training environment}

We used PyTorch implementation of ESPNetv2 \cite{Mehta2019}
with modifications of adding an auxiliary classification branch,
and normalizing the features and classification weights.
The same architecture was used in the baseline methods.
The models were trained and evaluated on
one NVIDIA Quadro RTX 8000 with 48GB RAM.
The network is trained with
an initial learning rate of $2\times 10^{-2}$
and cyclic learning rate scheduling \cite{Smith2017}.

\subsubsection{Datasets}

As source datasets, we use CamVid (CV) \cite{Brostow2009},
Cityscapes (CS) \cite{Cordts2016},
and Freiburg Forest (FR) \cite{Valada2017}.
As target datasets, we use \textit{Greenhouse A} \cite{Matsuzaki2022a};
\textit{TUT Park} \cite{Uzawa2023},
images around an in-campus park;
and \textit{Toyohashi Trail} \cite{Uzawa2023},
images in unstructured mountain paths in Toyohashi Nature Trail.
The datasets are summarized in Table \ref{table:datasets}.
\begin{table}[tb]
	\centering
	\scriptsize
	\caption{Datasets used in the experiments}
	\label{table:datasets}
	\vspace{-2pt}
	\begin{tabular}{ c c c c c }
		\toprule
		                           & Name                                   & Train & Test & Description         \\
		\midrule
		\multirow{3}[0]{*}{Source} & Camvid (CV) \cite{Brostow2009}         & 367   & 233  & Urban               \\
		                           & Cityscapes (CS) \cite{Cordts2016}      & 2970  & 500  & Urban               \\
		                           & Freiburg Forest (FR) \cite{Valada2017} & 230   & 136  & Unpaved outdoor     \\
		\midrule
		\multirow{3}[0]{*}{Target} & Greenhouse A \cite{Matsuzaki2022a}     & 6689  & 100  & Tomato greenhouse   \\
		                           & TUT Park     \cite{Uzawa2023}          & 859   & 10   & Pavements in campus \\
		                           & Toyohashi Trail \cite{Uzawa2023}       & 9137  & 10   & Unpaved mountain    \\
		\bottomrule
	\end{tabular}
\end{table}

\subsubsection{Baselines}

In our comparative studies, we employ baseline methods as follows.

\textbf{Training without DA}
As baselines without DA, we use two methods,
namely \textit{supervised} and \textit{ensemble}.
In \textit{supervised}, a segmentation model is trained with
the three source datasets with labels converted to
the common target label set using
the label conversions.
\textit{Ensemble} merges the predicted object probabilities
from the source models
in the same way as the proposed soft pseudo-label generation.
In other words, the soft pseudo-labels are directly used
as prediction results.

\textbf{Single-source DA}
We also evaluate single-source DA methods.
As source datasets, we use CV, CS, and FR in Greenhouse A,
and CS and FR in TUT Park dataset.
Pseudo-labels are generated from a model trained on each source
followed by label conversion $\psi_i$.

\textbf{Multi-source DA}
We use our previous method \cite{Matsuzaki2022a} as a baseline
of multi-source DA, referred to as \textit{MSPL}.
We also use the method by He et al. \cite{He2021},
a state-of-the-art MDA for semantic segmentation,
hereafter referred to as \textit{MSDA\_CL}.
We used our own implementation of the method \cite{He2021}.
In multi-source DA, we evaluate
both double-source and triple-source settings.
For double-source training,
we use CS and FR as source datasets.

%

\subsection{Comparisons with the baselines}


\begin{table}[tb]
	\centering
	\caption{Comparison with the baselines on TUT Greenhouse A}
	\label{table:comparison_with_baselines_greenhouse}
	\vspace{-2pt}
	\scriptsize
	\begin{tabular}{cccccc}
		\toprule
		                          & \multirow{2}{*}[-1.5em]{{Method}} & \multicolumn{3}{c}{{Class IoU}} & \multirow{2}{*}[-1.5em]{{mIoU}}                                             \\
		\cmidrule{3-5}
		                          &                                   & \rotatebox{90}{Plant}           & \rotatebox{90}{Artificial}      & \rotatebox{90}{Ground}                    \\
		\midrule
		\multirow{2}{*}{No adapt} & Supervised                        & 75.4                            & 64.8                            & 56.7                   & 65.6             \\
		                          & Ensemble                          & 74.8                            & 70.5                            & 50.8                   & 65.4             \\
		\midrule
		                          & CV                                & 71.0                            & 62.7                            & 50.7                   & 61.5             \\
		Single                    & CS                                & 65.0                            & 72.9                            & 40.8                   & 59.6             \\
		                          & FR                                & 67.2                            & 52.9                            & 37.8                   & 52.7             \\
		\midrule

		                          & MSDA\_CL \cite{He2021}            & 70.9                            & 57.9                            & 52.8                   & 60.5             \\
		Double                    & MSPL \cite{Matsuzaki2022a}        & \underline{80.7}                & 74.9                            & \underline{67.6}       & 74.4             \\
		                          & \textbf{Proposed}                 & 79.9                            & 76.4                            & 57.4                   & 71.2             \\
		\midrule
		                          & MSDA\_CL \cite{He2021}            & 70.5                            & 62.1                            & 43.9                   & 58.9             \\
		Triple                    & MSPL \cite{Matsuzaki2022a}        & \underline{80.7}                & \textbf{78.2}                   & \textbf{72.6}          & \textbf{77.2}    \\
		                          & \textbf{Proposed}                 & \textbf{82.7}                   & \underline{77.0}                & 64.7                   & \underline{74.8} \\
		\bottomrule
	\end{tabular}
\end{table}

\textbf{Greenhouse A dataset}
Table \ref{table:comparison_with_baselines_greenhouse} shows
the results of the baselines and the proposed method.
The performance of the proposed method did not
reach those of our previous method (MSPL).
However, the proposed method resulted in the second-best mean IoU
and outperformed MSDA\_CL \cite{He2021}.
In MSDA\_CL, pseudo-labels are generated by simply adding predicted scores
from the source models, followed by
confidence-based label selection \cite{Zou2018}.
In contrast, the proposed method considers domain similarity between the source datasets and the target.
Moreover, by loss weighting using the label entropy, the effect of noisy
pseudo-labels is suppressed, resulting in better performance.
We further evaluate the effect of considering domain similarity in pseudo-label generation
and label entropy weight in training in \ref{sec:ablation}.

\begin{table}[tb]
	\centering
	\vspace{-5pt}
	\caption{Comparison with the baselines on TUT Park}
	\label{table:comparison_with_baselines_sakaki}
	\vspace{-2pt}
	\scriptsize
	\begin{tabular}{cccccccc}
		\toprule
		                          & \multirow{2}{*}[-1.5em]{{Method}} & \multicolumn{5}{c}{{Class IoU}} & \multirow{2}{*}[-1.5em]{{mIoU}}                                                                                                \\
		\cmidrule{3-7}
		                          &                                   & \rotatebox{90}{Plant}           & \rotatebox{90}{Grass}           & \rotatebox{90}{Artificial} & \rotatebox{90}{Ground} & \rotatebox{90}{Sky} &                  \\
		\midrule
		\multirow{2}{*}{No adapt} & Supervised                        & 79.8                            & 47.9                            & 61.8                       & 35.0                   & 78.7                & 60.6             \\
		                          & Ensemble                          & 82.8                            & 37.5                            & \textbf{68.2}              & 23.9                   & 77.3                & 57.9             \\
		\midrule
		\multirow{2}{*}{Single}   & CS                                & 82.2                            & 22.4                            & 59.5                       & 17.7                   & \textbf{90.4}       & 54.4             \\
		                          & FR                                & 74.8                            & 71.2                            & 8.2                        & \textbf{57.1}          & 55.6                & 53.4             \\
		\midrule

		                          & MSDA\_CL \cite{He2021}            & 81.6                            & 60.3                            & 43.9                       & 33.8                   & 72.8                & 58.5             \\
		Double                    & MSPL \cite{Matsuzaki2022a}        & 79.7                            & \underline{65.3}                & 45.4                       & \underline{44.2}       & 76.9                & 62.3             \\
		                          & \textbf{Proposed}                 & 73.7                            & \textbf{67.4}                   & 48.9                       & 51.4                   & {86.5}              & \underline{65.6} \\
		\midrule
		                          & MSDA\_CL \cite{He2021}            & \textbf{85.8}                   & 0.67                            & \underline{67.4}           & 15.2                   & 82.0                & 50.2             \\
		Triple                    & MSPL \cite{Matsuzaki2022a}        & \underline{84.0}                & 0.00                            & 51.6                       & 16.4                   & 81.0                & 46.6             \\
		                          & \textbf{Proposed}                 & 83.5                            & 62.0                            & 58.9                       & 37.2                   & \underline{88.2}    & \textbf{65.9}    \\
		\bottomrule
	\end{tabular}
\end{table}

\textbf{TUT Park dataset}
In this target dataset, we use five object classes that
dominate the scenes,
namely \textit{plant} (e.g., trees), \textit{grass} (e.g., ground vegetation),
\textit{artificial object}, \textit{road}, and \textit{sky}.
CS and FR have a distinction between \textit{plant} (\textit{vegetation}
in CS, and \textit{tree} in FR) and \textit{grass}
(\textit{terrain} in CS, and \textit{grass} in FR).
CamVid, however, only has \textit{tree} class
that corresponds to \textit{plant} in the target dataset.

Table \ref{table:comparison_with_baselines_sakaki} shows the results.
Interestingly, the proposed method
trained with three sources did not perform the best on
any classes, but resulted in the best mean IoU.
While the baseline methods were biased towards a specific class
and poorly performed on others,
the proposed method realized well-balanced training.
Notably, while MSPL by nature failed to capture
the \textit{grass} class due to its absence in one of the source datasets (CV),
the proposed method successfully learned it.
Unlike the previous method which excludes from pseudo-labels
the pixels on which the source models did not agree with each other,
our soft pseudo-labeling allowed for learning
classes induced by only part of source datasets.

\begin{table}[tb]
	\centering
	\caption{Comparison with the baselines on Toyohashi Trail}
	\label{table:comparison_with_baselines_imo}
	\vspace{-2pt}
	\scriptsize
	\begin{tabular}{cccccccc}
		\toprule
		                          & \multirow{2}{*}[-1.5em]{{Method}} & \multicolumn{5}{c}{{Class IoU}} & \multirow{2}{*}[-1.5em]{{mIoU}}                                                                                                \\
		\cmidrule{3-7}
		                          &                                   & \rotatebox{90}{Plant}           & \rotatebox{90}{Grass}           & \rotatebox{90}{Artificial} & \rotatebox{90}{Ground} & \rotatebox{90}{Sky} &                  \\
		\midrule
		\multirow{2}{*}{No adapt} & Supervised                        & 58.3                            & 2.05                            & 17.7                       & 15.9                   & 0.0                 & 18.8             \\
		                          & Ensemble                          & 69.0                            & 1.34                            & 0.14                       & 0.28                   & 0.0                 & 14.1             \\
		\midrule
		\multirow{2}{*}{Single}   & CS                                & 65.2                            & 3.67                            & 2.97                       & 32.6                   & \textbf{63.1}       & {33.5}           \\
		                          & FR                                & 63.8                            & 24.8                            & \textbf{34.3}              & 22.5                   & 0.42                & 29.2             \\
		\midrule

		                          & MSDA\_CL \cite{He2021}            & \textbf{74.6}                   & \underline{30.5}                & 4.10                       & 56.9                   & 3.25                & \underline{34.0} \\
		Double                    & MSPL \cite{Matsuzaki2022a}        & 70.4                            & 9.36                            & 14.7                       & 50.2                   & 0.75                & 29.1             \\
		                          & \textbf{Proposed}                 & 68.9                            & 24.4                            & 0.0                        & \underline{59.4}       & 0.16                & 30.6             \\
		\midrule
		                          & MSDA\_CL \cite{He2021}            & \underline{72.2}                & 0.01                            & \underline{28.1}           & 46.7                   & 1.99                & 29.8             \\
		Triple                    & MSPL \cite{Matsuzaki2022a}        & 68.0                            & 0.00                            & 19.7                       & 53.4                   & \underline{17.5}    & 31.7             \\
		                          & \textbf{Proposed}                 & 71.3                            & \textbf{33.4}                   & 19.6                       & \textbf{59.7}          & 1.06                & \textbf{37.0}    \\
		\bottomrule
	\end{tabular}
\end{table}

\textbf{Toyohashi Trail dataset}
We use the same label set as TUT Park dataset.
Table \ref{table:comparison_with_baselines_imo} shows the comparative results.
The proposed method resulted in the best mean IoU.
Although the domain gap estimation was highly unreliable due to
large domain gap between the all source datasets and the target datasets,
and thus it is difficult to give theoretical justification to
the effect of considering domain similarity,
they show a better capability of the proposed method
to transfer knowledge even from source datasets that are
very different from the target dataset compared to the baselines.
%

\subsection{Ablation studies}
\label{sec:ablation}

Next, we conducted an ablation study on
loss weighting based on entropy of the pseudo-labels during training,
and the domain similarity-based integration of source predictions in
pseudo-label generation.
We used TUT Park as the target dataset,
and trained the network using CV, CS, and FR.

\begin{figure}[tb]
	\centering

	\subfloat[Input]{
		\includegraphics[width=0.44\hsize]{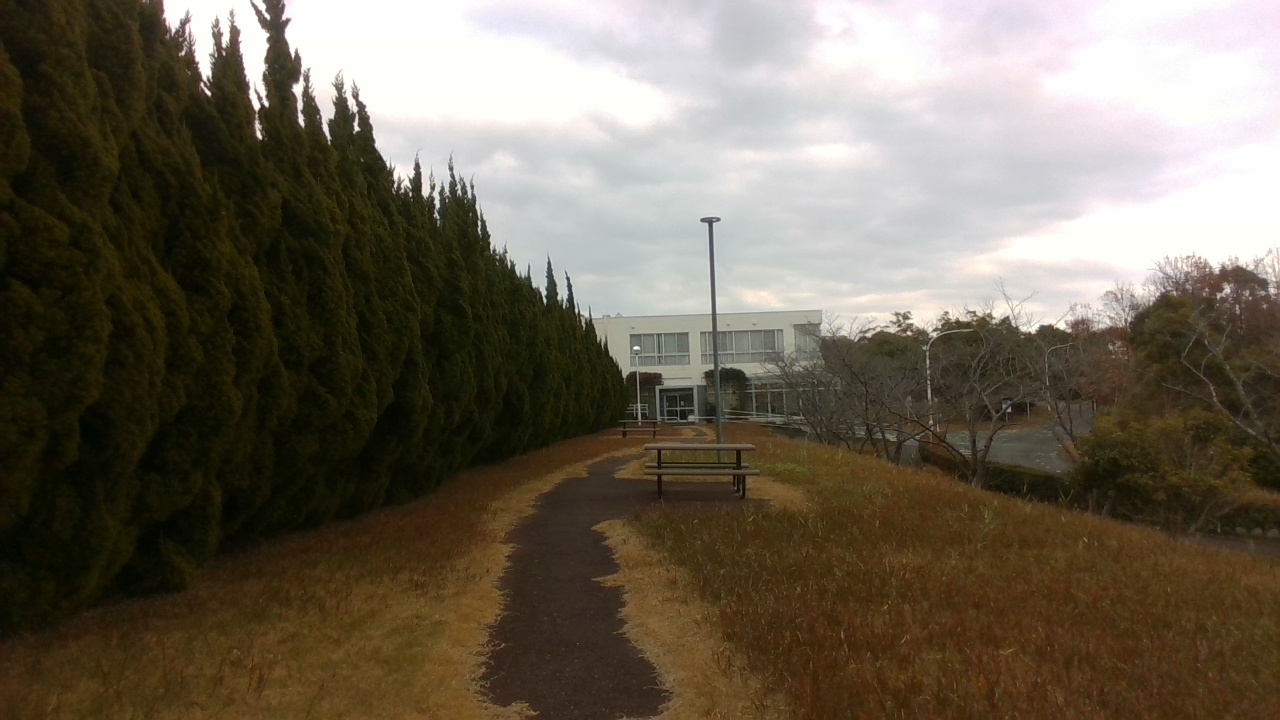}
		\label{fig:ablation_pseudo-label_input}}
	\subfloat[Ground truth]{
		\includegraphics[width=0.44\hsize]{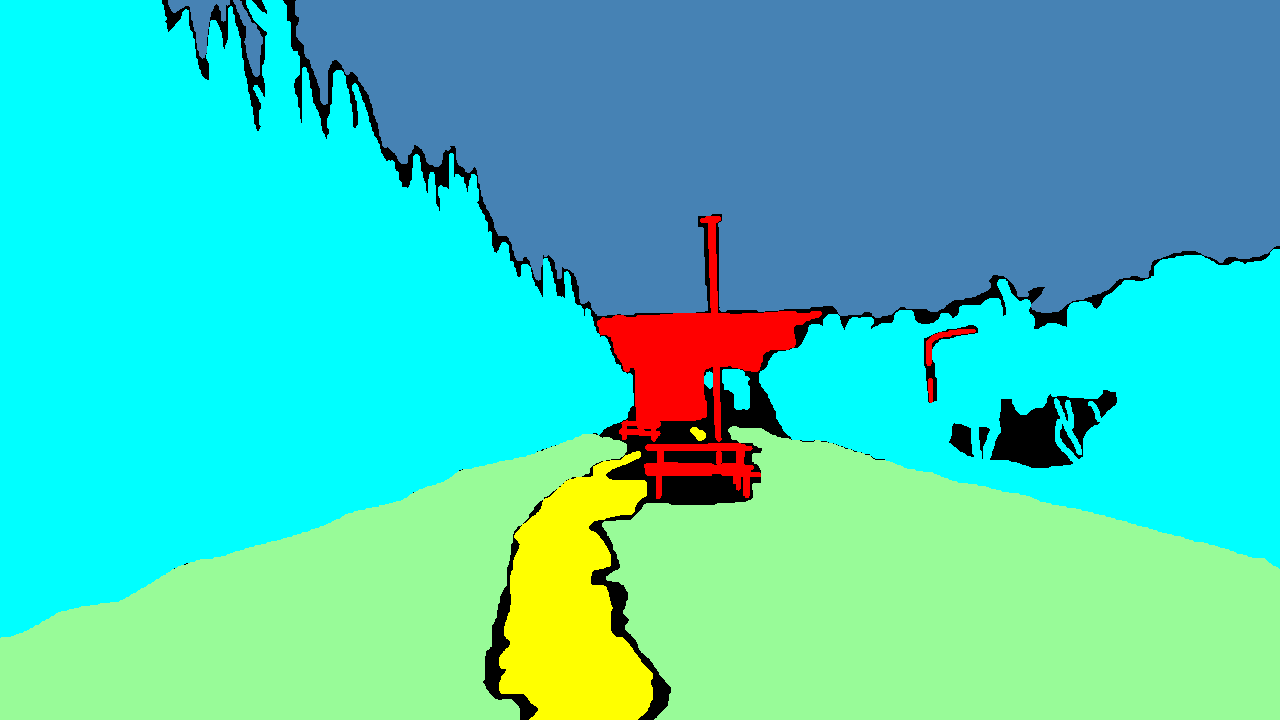}
		\label{fig:ablation_pseudo-label_gt}}

	\subfloat[w/o domain similarity]{
		\includegraphics[width=0.44\hsize]{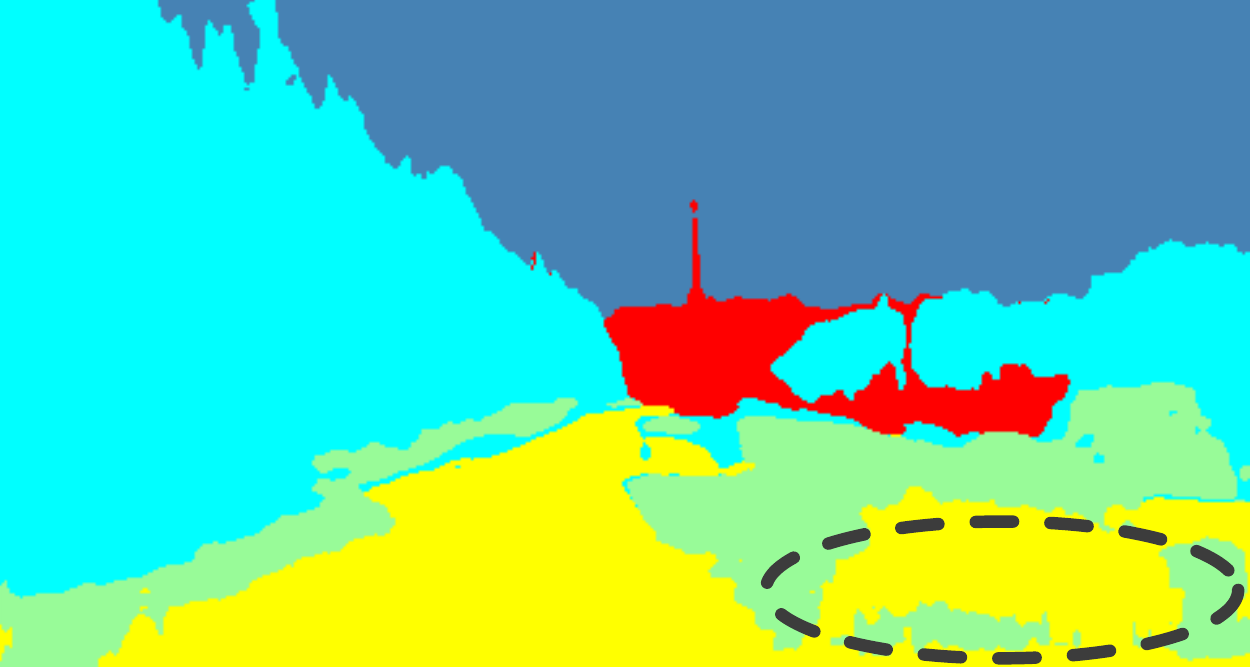}
		\label{fig:ablation_pseudo-label_none}}
	\subfloat[w/ domain similarity]{
		\includegraphics[width=0.44\hsize]{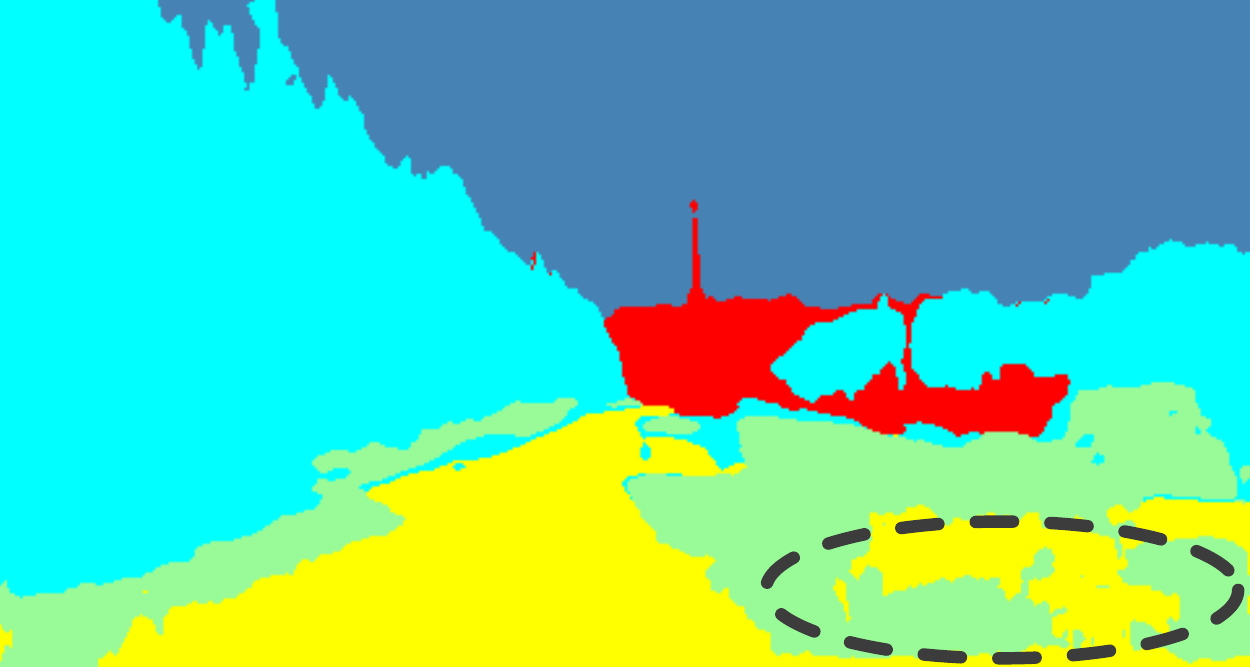}
		\label{fig:ablation_pseudo-label_per_sample}}
	\caption{Comparison of pseudo-label generated with and without considering domain similarity.
		(c) When domain similarity is not incorporated,
		a large area of \textit{grass} is misclassified as \textit{ground}.
		(d) Some pixels misclassified in (c) are classified correctly
		(inside the dotted ellipse).
	}
	\label{fig:ablation}
\end{figure}

Table \ref{table:ablation} shows the results.
Using domain similarity in pseudo-label generation
improved the performance in both
training settings with and without the label entropy weights.
Fig. \ref{fig:ablation} shows an example of
pseudo-label generated with and without domain similarity weighting.
While a large area of \textit{grass} is wrongly assigned
with \textit{ground} class when source predictions are simply summed
as shown in Fig. \ref{fig:ablation_pseudo-label_none},
part of the area is correctly classified when domain similarity is incorporated
as shown in Fig. \ref{fig:ablation_pseudo-label_none}.
By considering domain similarity,
the prediction of a source model with larger domain similarity was
emphasized and pseudo-labels becomes more accurate.
The effect of using domain similarity is more evident when
label entropy weight is not employed in training
because the improvement of pseudo-labels occurs on pixels
where the source predictions are fuzzy,
which are suppressed in training by label entropy weight.
We conclude that incorporating domain similarity
brings about positive effect on pseudo-label generation.
\begin{table}[tb]
	\centering
	\caption{Ablation study on training method}
	\label{table:ablation}
	\vspace{-2pt}
	\begin{tabular}{ c c c }
		\toprule
		Label entropy weight & Domain similarity & mIoU          \\
		\cmidrule(lr){1-2} \cmidrule{3-3}
		                     &                   & 60.7          \\
		                     & \checkmark        & 62.3          \\
		\checkmark           &                   & 65.8          \\
		\checkmark           & \checkmark        & \textbf{65.9} \\
		\bottomrule
	\end{tabular}
\end{table}

\section{Conclusion}

We proposed a method of
soft pseudo-label generation
for training semantic segmentation models
on datasets of a variety of scenes
without ground-truth labels
using not very relevant source datasets.
Unlike our previous method \cite{Matsuzaki2022a}
using the unanimity criterion for pseudo-label selection,
our method allows for
taking into account the domain similarity between each source dataset
and the target dataset,
utilizing the information of the prediction certainty
of the source models,
and involving target classes that does not have a corresponding
class in some source datasets,
which increase the applicability of the method to a variety of scenes.




%




\printbibliography

\end{document}